\documentclass[a4paper,twoside]{article}

\usepackage{times}
\usepackage{epsfig}
\usepackage{graphicx}
\usepackage{amsmath}
\usepackage{amssymb}
\usepackage[utf8]{inputenc}
\usepackage{pdfpages}
\usepackage{caption}
\usepackage{multirow}
\usepackage{url}
\usepackage{hyperref}
\usepackage{apalike}
\usepackage{textcomp}
\usepackage{SCITEPRESS}     

\begin{document}

\title{Towards Keypoint Guided Self-Supervised Depth Estimation}

\author{%
  \authorname{%
    Kristijan Bartol\sup{1},
    David Bojani{\'c}\sup{1},
    Tomislav Petkovi{\'c}\sup{1},
    Tomislav Pribani{\'c}\sup{1},
    Yago Diez Donoso\sup{2}%
  }%
  \affiliation{\sup{1}University of Zagreb, Faculty of Electrical Engineering and Computing, Zagreb, Croatia, EU}%
  \affiliation{\sup{2}Yamagata University, Faculty of Science, Yamagata, Japan}%
  \email{\{kristijan.bartol, david.bojanic, tomislav.petkovic.jr, tomislav.pribanic\}@fer.hr, yago@sci.kj.yamagata-u.ac.jp}
}%



\abstract{%
This paper proposes to use keypoints as a self-supervision clue for learning depth map estimation from a collection of input images. As ground truth depth from real images is difficult to obtain, there are many unsupervised and self-supervised approaches to depth estimation that have been proposed. Most of these unsupervised approaches use depth map and ego-motion estimations to reproject the pixels from the current image into the adjacent image from the image collection. Depth and ego-motion estimations are evaluated based on pixel intensity differences between the correspondent original and reprojected pixels. Instead of reprojecting the individual pixels, we propose to first select image keypoints in both images and then reproject and compare the correspondent keypoints of the two images. The keypoints should describe the distinctive image features well. By learning a deep model with and without the keypoint extraction technique, we show that using the keypoints improve the depth estimation learning. We also propose some future directions for keypoint-guided learning of structure-from-motion problems.
}

\onecolumn \maketitle \normalsize \setcounter{footnote}{0} \vfill

\section{\uppercase{Introduction}}

\begin{figure*}[h!]
  \centering
  \captionsetup{justification=centering}
  \includegraphics[width=0.9\linewidth]{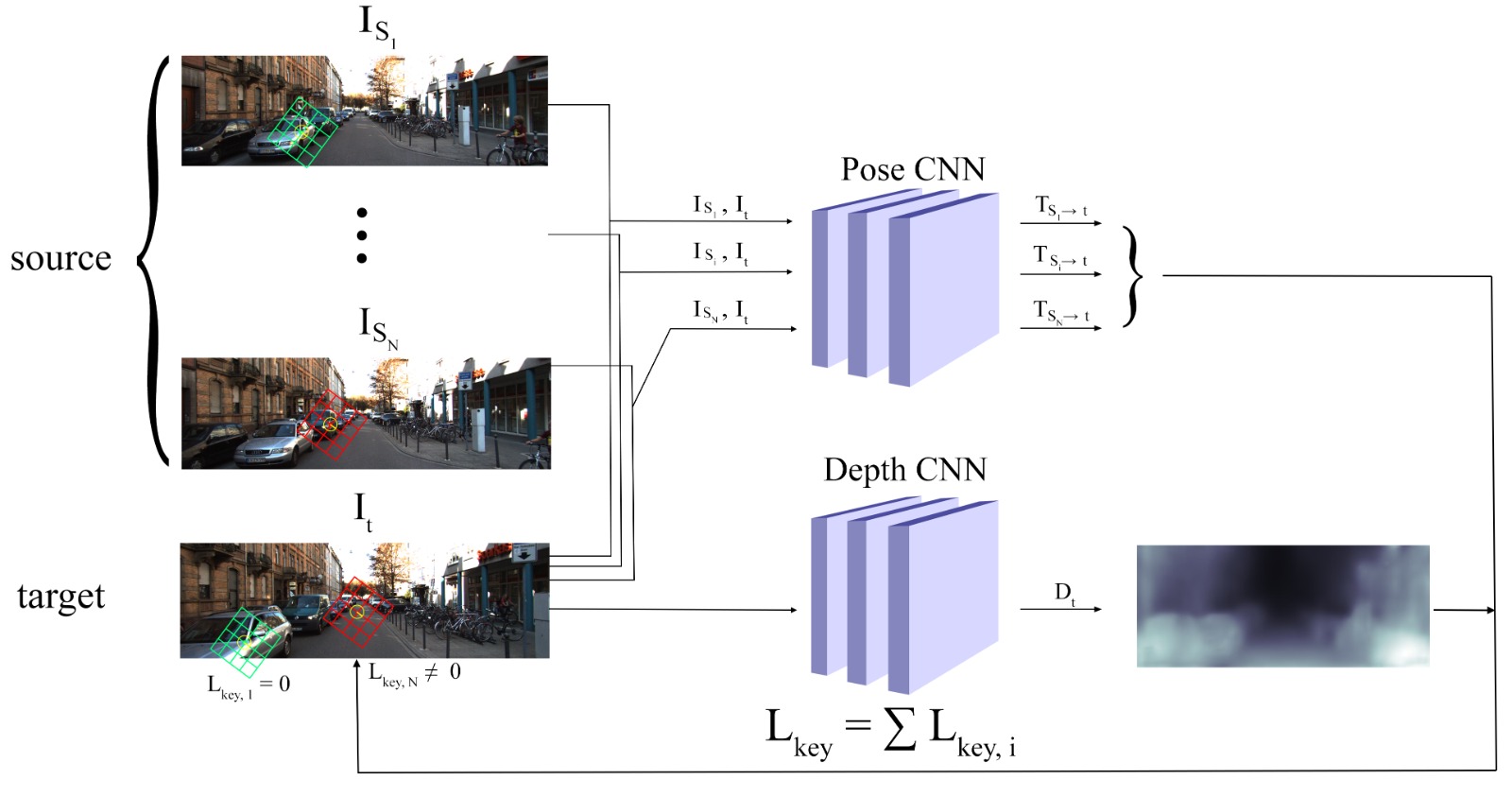}
  \caption{Overview of the model. The model contains separate depth and pose networks that jointly learn as in \cite{garg} and \cite{sfm-learner}. Instead of source images' pixels, the keypoints of the source images are reprojected to the target image. The keypoints shown in green is the correct reprojection and the keypoints in red is the incorrect one. Note that the reprojection is done for all the source-target image pairs separately.}
  \label{fig:overview}
\end{figure*}

Monocular depth estimation is a long-standing, ill-posed computer vision problem. A depth map estimated from a monocular image describes an infinite amount of scenes due to the scale ambiguity. Nevertheless, monocular depth estimation is a very popular topic, especially in the deep learning era. A particularly interesting approach is the joint, unsupervised learning of monocular depth and pose \cite{garg}. The model has two convolutional networks, one which outputs a depth map and one which outputs the transformation matrices representing pose transformations between the target and the source views (\autoref{fig:overview}). Assuming the intrinsic matrix is known, depth and pose estimations are sufficient to reproject the pixels from the source views to the target view \cite{zisserman}. The sum of differences between the original (target) and the reprojected (source) pixel intensities is called a photometric loss.

The supervision clue in case of unsupervised learning comes from time component between the images in a collection. The idea of the photometric loss is to learn to warp the source image to match with the target image. Another way to look at this is that the photometric loss is used to learn the model to find the correct pixel correspondences between the images. Of course, pixel intensities are not unique, so even though the correspondent pixel intensities are the same, it does not guarantee that they are truly correspondent. For example, the pixels of the white wall will perfectly match, even though they might not be correspondent in 3D. 

There are many ways to cope with the correspondence problem. Unsupervised depth estimation models like \cite{sfm-learner} estimate depth maps on multiple scales. The pixel on a lower scale is aggregated from the square of pixels on the original scale. It is therefore expected that the lower scale pixels will match better in case of low or repeating textures. In classical structure-from-motion, for example, in COLMAP \cite{colmap}, the correspondences are found by matching the selected image keypoints. Our proposal is therefore to select and reproject the keypoints from the source to the target images and then compare these keypoints to determine their similarity. If the values are similar, it means that the keypoints of both images are probably representing the same part of the 3D scene, so the loss function value should be low, and vice versa. To the best of our knowledge, the keypoints have not yet been utilitized for learning structure-from-motion.

We use SIFT keypoint descriptors by \cite{sift} to compare the original and the reconstructed image. Keypoints have many beneficial properties. First, they preserve and enforce distinctiveness of image regions. In \cite{deep-vs-classic-keypoints}, it is shown that the SIFT descriptors are still among the best options in state-of-the-art of keypoint descriptors. Second, the selected keypoints are expected to be more important and informative than other image regions. Third, SIFT keypoints are assigned varying sizes that are, in general, reversely proportional to the potential information in the pixel neighbourhood. For example, low texture region might be assigned a large sized keypoint whose boundary reaches some distinctive edges, also making this low texture region more distinctive, carrying greater information. Large sized keypoint regions offer an elegant solution to handling low or repeating texture areas compared to multi-scale depth estimation. Finally, by selecting the keypoints, the model also ignores the regions that are not beneficial, for example, very large textureless areas like sky, road or walls. Learning models like \cite{sfm-learner} cope with this by learning the explainability mask that assigns weights to each pixel based on their estimated importance.

The aim of this work is to show that using the keypoints provide a beneficial clue for self-supervised learning of depth estimation. To summarize, we propose a \textbf{keypoint similarity loss} between the original and the reprojected image keypoints as an improvement to the unsupervised loss components' stack and as a replacement for the explainability mask loss and multi-scale depth map estimation.

\section{\uppercase{Related Work}}
    \label{section:related-work}

There is a lot of work dedicated to depth estimation. In this section we will give a brief overview over the recent attempts which are mostly focused on deep learning.

\begin{figure*}[h!]
  \centering
  \captionsetup{justification=centering}
	\includegraphics[width=0.8\linewidth]{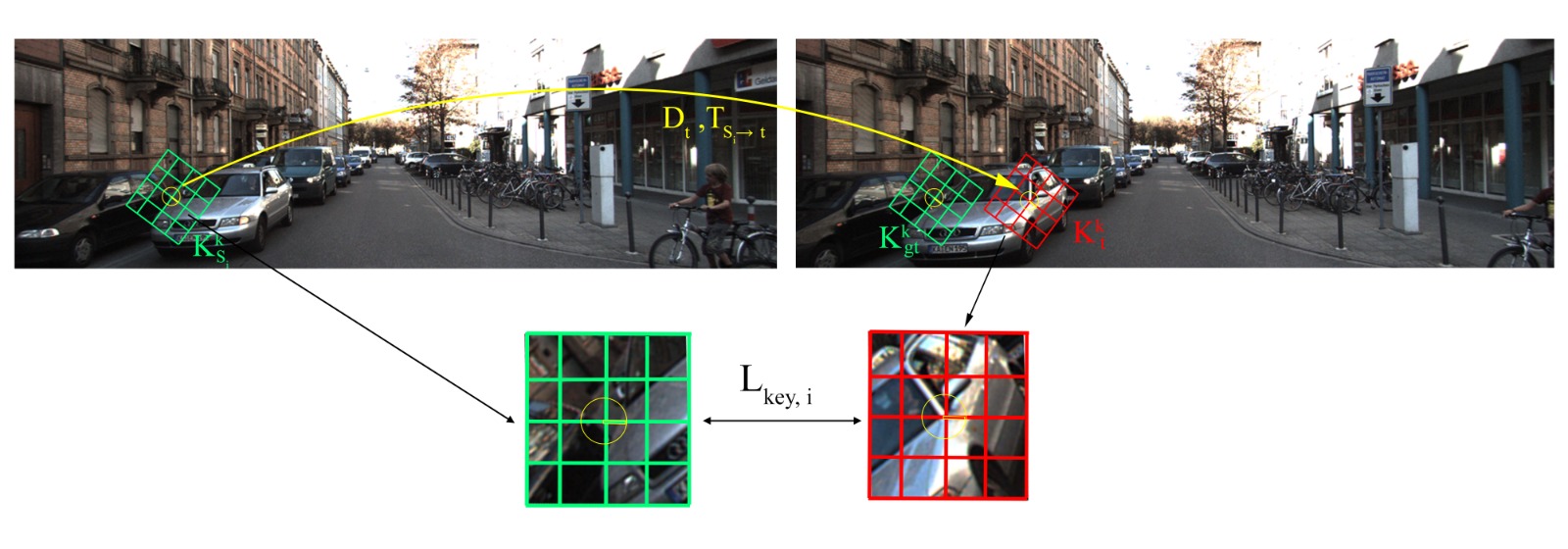}
	\caption{The keypoint similarity loss. The selected keypoint on the left (source) image is reprojected to the right (target) image using the depth and pose estimations. The red keypoint indicates that the reprojection is incorrect. The difference between the keypoint descriptors is labeled as $L_{key, i}$ and is a keypoint similarity loss for a given pair of keypoints.}
	\label{fig:keypoint-similarity}
\end{figure*}

\textbf{Unsupervised structure-from-motion.} The pioneering unsupervised learning of monocular depth estimation work is done by \cite{garg}. The authors propose an image warping technique, the same as the one used in this work. They also propose a smooth loss function that minimizes the differences between the neighbouring values of an estimated depth map. It is shown that smoothing the depth map greatly improves the estimation accuracy and serves as a regularization for the photometric loss. Instead of using time as a supervision clue, they use a stereo pair and compare between the real and the reconstructed image. A similar approach is proposed by \cite{monodepth} to enable left-right consistency check.

A model by \cite{sfm-learner} learns to predict both the depth map and pose estimations between the views, exploiting the time component, as done in this paper. The photometric loss uses the image warping technique and compares the target image with its reconstructions sampled from the source images. The model architecture is composed of depth and pose estimation networks which are coupled during training as shown in \autoref{fig:overview}, but which can be applied independently in test time. They also propose to learn the explainability mask whose goal is learning to ignore the parts of the image that might degrade the photometric loss performance. For example, occluded or moving objects are ideally ignored using the explainability mask. Finally, they output depth maps on multiple scales to enable pixels of smaller scale to see larger patches of the original image and in that way cope with low textured regions.

The authors of \cite{monodepth2} improve the depth estimation results by focusing on the pose estimation network. The authors propose to share the encoder weights between the depth and pose network. Also, they use the improved, edge-aware smooth loss that accounts less for the differences between the neighbouring depth map values if their corresponding, original image difference is also higher. Instead of directly utilizing the depth maps on smaller scale, they simply upscale these depth maps to the original image size and then apply the photometric error, which is shown to reduce the texture-copy artifacts. That way they improve on the multi-scale depth map estimation.

\textbf{Keypoint similarity.} LF-Net \cite{lf-net} is a self-supervised model for learning keypoint detection and description. Similar to us, they also use SIFT keypoints. On top of the keypoints supervision, SfM algorithm is used to estimate the transformations between the image pairs so that the LF-Net model is not directly supervised by SIFT (otherwise, it would perform as SIFT at best). The model predicts the keypoints for the reference image and then these keypoints are transformed to the ground truth image where the detections and the corresponding descriptions are compared. The difference between LF-Net and our proposal is that LF-Net uses SIFT and SfM self-supervision to learn to generate keypoints. Our model directly uses SIFT keypoints to learn SfM, i.e., the keypoints help the model to verify the keypoint correspondence, which is the core problem in SfM \cite{mvs-tutorial}. We further reflect on LF-Net in \autoref{section:future-work}.

\section{\uppercase{Keypoint similarity loss}}
    \label{section:keypoint-similarity-loss}
    
\begin{figure*}[h!]
  \centering
  \captionsetup{justification=centering}
	\includegraphics[width=0.8\linewidth]{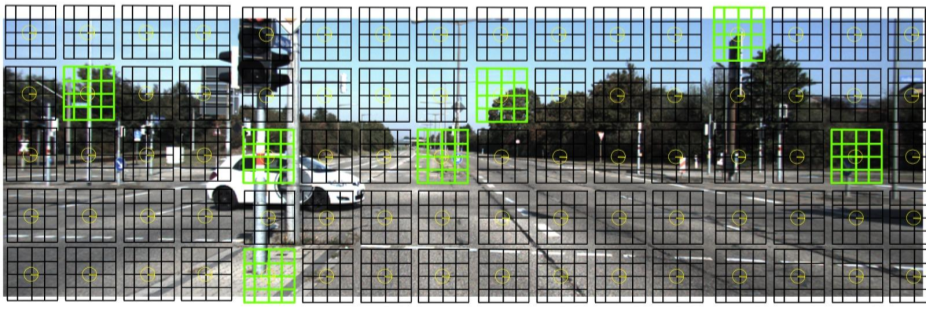}
	\caption{The keypoints are precomputed in every pixel location. The green keypoints represent the keypoints selected by the detector. The keypoints have predefined sizes and orientations.}
	\label{fig:sift-grid}
\end{figure*}    

Let $I_t$ denote a target image, $I_{s_i}$ one of the source images, $D_t$ a depth map estimated for the target image, matrix $K$ the camera intrinsics and $T_{s_i \to t}$ a rigid transformation between the views (pose). The standard photometric loss evaluates the depth estimation of the target view, $D_t$, by measuring how well the pixels from the source image reproject to the target image. Precisely, $I_t$ is reconstructed by warping $I_{s_i}$ based on $D_t$ and $T_{s_i \to t}$:

\begin{equation}
\label{eq:photometric_loss}
    \hat{I}_t^{ij} = I_{s_i}^{uv} = I_{s_i}^{u^{*} v^{*}} = K T_{s_i \to t} (D_t^{ij} K^{-1} I_{s_i}^{ij}),
\end{equation}

where $\hat{I}_t^{ij}$ is the reconstructed target image sampled from the image $I_{s_i}$ in the coordinates $(u, v) \to I_{s_i}^{uv}$. Note that the pixel $(i, j)$ reprojects to the subpixel $(u^{*}, v^{*})$. To assign the exact $(u, v)$ pixel's intensity, the $(u^{*}, v^{*})$ subpixel reprojection is used to sample the four closest pixel intensities using bilinear interpolation. The difference between the pixel intensities of the original image $I_t$ and of the reconstructed image $\hat{I}_t$ is the photometric loss $L_{photo} = \sum_{ij} ||\hat{I}_t^{ij} - I_t^{ij}||_1$. Instead of reprojecting and comparing all the pixel intensities, we propose to only reproject and compare the keypoints selected by SIFT. The overview of the proposal is shown in \autoref{fig:overview}.

Let $K_{s_i}^{k}$ denote $k$-th keypoint in the source image and $K_{t}^{k}$ its correspondent keypoint in the target image (\autoref{fig:keypoint-similarity}). We define the keypoint similarity loss function between the two images $I_{t}$ and $I_{s_i}$ as a sum over the differences between the correspondent keypoint vectors:

\begin{equation}
\label{eq:key_sim}
    L_{key, i} = \sum_k \sum_l ||K^{k}_t (l) - K^{k}_{s_i} (l)||_1,
\end{equation}

where $k$ represents $k$-th keypoint of the source image $I_{s_i}$ and $l \in {1...128}$ the index of an element in the $k$-th SIFT keypoint vector. For the source keypoint $K_{s_i}$ and its perfectly reprojected, corresponding counterpart $K_{gt}^k$, the loss should be zero (\autoref{fig:keypoint-similarity}). The total keypoint similarity loss is the sum of keypoint similarity losses for all the source-target image pairs, $L_{key} = \sum_i L_{key, i}$.

For the experimental purposes, we also use other loss components previously proposed and used by \cite{sfm-learner}, \cite{monodepth2}, etc., for example, smooth loss and explainability mask loss. When all these components are included, our loss function looks like:

\begin{equation}
    \label{eq:total_loss}
    L = \alpha L_{key} + \beta L_{photo} + \gamma L_{smooth} + \delta L_{expl},
\end{equation}

where $\alpha$, $\beta$, $\gamma$ and $\delta$ are the loss hyperparameters that balance the dynamics between the loss components. The best hyperparameter values determined by the experiments are: $\alpha = 2.0$, $\beta = 1.0$, $\gamma = 0.5$ and $\delta = 0.2$.

\section{\uppercase{Experiments}}
    \label{section:experiments}

The purpose of the experiments is to provide a step towards the keypoint guided self-supervision for learning depth estimation. For this purpose, we choose the KITTI dataset \cite{kitti} and the model based on \cite{sfm-learner}, but the similar approach can be applied to other structure-from-motion environments and datasets. Instead of using the original smooth loss, we use the improved, edge-aware smoothness from \cite{monodepth}. To enable the keypoint similarity loss, we add it as a loss component, based on the Eq. \ref{eq:key_sim} from \autoref{section:keypoint-similarity-loss}. 

The SIFT keypoints are precomputed in every pixel (\autoref{fig:sift-grid}), for two reasons. First, by precomputing the keypoints in all the pixel locations $(i, j)$, we make sure that the keypoint reprojections in $(u, v)$ will have its correspondent pair, for every $(u, v)$ inside the image boundaries. Second, by precomputing the keypoints in every pixel, we are able to test two different approaches - learning with and without the keypoint \textit{selection} mechanism. In case learning is done without the keypoint detector, keypoints calculated from every source image pixel's neighbourhood reproject to the target image, which boils down to using keypoint descriptions instead of pixel intensities as a supervision clue. 

When using the keypoint detector, the gradients in backward pass are applied only on the selected keypoints and the rest are masked as shown in \autoref{fig:sift-grid}. As expected, we show that keypoint detection improves learning, compared to the model that does not select the keypoints. We explain this by the fact that the SIFT descriptions of the keypoints are simply better for the selected keypoints, but it also indicates that the selectivity increases the quality of the backward passed gradients. To compute the keypoint for a specific pixel location $(i, j)$, the orientation and patch size need to be specified upfront by hand. We choose to provide zero angled orientations and the patches of size 15x15.

\begin{table*}[t]
\captionsetup{justification=centering}
\begin{tabular}{|c|c|c|c|c|c|c|c|}
\hline
Experiment / Metric & Abs Rel & Sq Rel & $\delta < 1.25$ & $\delta < 1.25^2$  & $\delta < 1.25^3$ \\
\hline
\hline
No det + expl (ours) & 0.309 & 3.813 & 0.603 & 0.818 & 0.912 \\
\hline
Det + expl (ours) & 0.284 & 3.275 & 0.620 & 0.822 & 0.914 \\
\hline
No det + no expl (ours) & 0.282 & 3.160 & 0.605 & 0.815 & 0.911 \\
\hline
Det + no expl (ours) & \textbf{0.277} & 3.275 & \textbf{0.633} & \textbf{0.829} & \textbf{0.917} \\
\hline
\hline
SfMLearner & 0.286 & \textbf{3.072} & 0.629 & \textbf{0.829} & 0.916 \\
\hline
SfMLearner (full) & \textit{0.181} & \textit{1.341} & \textit{0.733} & \textit{0.901} & \textit{0.964} \\
\hline
\end{tabular}
\centering
\caption{\label{tab:quantitative}Quantitative comparison of the experimental results. The first column displays the experiments' names. The second and third show absolute relative and squared relative error (the lower the better). The last three columns show the accuracy metrics, where $\delta$ denotes the ratio between the estimates and the ground truth (the higher the better). The first part of the table shows the results obtained when keypoint similarity loss was used.}
\end{table*}

\subsection{Quantitative Comparison}

The first row of the \autoref{tab:quantitative} shows the used metrics' names. The first part of the table shows the results of four of our different experiments and the second part shows the results of the base model (without the keypoint similarity loss). The four experiments are done by learning the model with and without the keypoint detector and with and without the explainability mask.

Looking at the first part of the \autoref{tab:quantitative}, it is shown that the model learn better without the explainability mask. As expected, the more selective models (the ones in third and fifth row of the table) are shown to learn better than their non-selective counterparts (second and fourth row). Our selective keypoint model without the explainability mask slightly outperforms SfMLearner. This is an indication that keypoint guidance help to improve the overall learning performance. The last row shows the results reported by \cite{sfm-learner} after training on KITTI and fine-tuning on Cityscapes dataset \cite{cityscapes}. Even though we outperform the base model using the keypoint similarity loss, the model needs to be further fine-tuned to reach the results in the last row of \autoref{tab:quantitative}.

\subsection{Qualitative Comparison}

\begin{figure*}[h!]
  \centering
  \captionsetup{justification=centering}
	\includegraphics[width=1\linewidth]{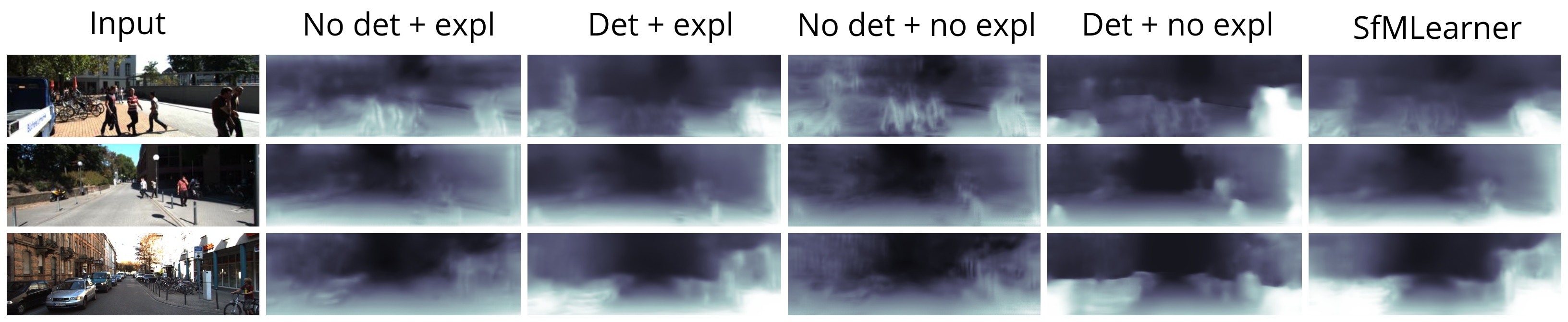}
	\caption{The overview of the qualitative results. The three input images (in the first column) are randomly selected from the KITTI dataset.}
	\label{fig:qualitative}
\end{figure*}

The overview of the qualitative results is shown in the \autoref{fig:qualitative} where the first column shows the input images and the rest of the columns follow the second to sixth row of the \autoref{tab:quantitative}. The fifth column contains the sharpest depth maps which corresponds with the quantitative results. Interestingly, the three people are most precisely reconstructed in the fourth row, when the keypoint selection was not used. Comparing the depth maps the models using the explainability mask with the ones not using it (first part of the qualitative results with the second), we confirm the quantitative results. The results from the models not using explainability mask are generally sharper and more precise and than the ones using the explainability. Finally, the last column is worse than the fifth column, both in terms of precision and sharpness. Moreover, the three people are not clearly visible at all in the last column.

\subsection{Implementation Details}

The model is an extension of the PyTorch \cite{pytorch} implementation of SfMLearner by \cite{sfm-learner}. Our experiments were run on a single NVidia GeForce RTX 2080 GPU, 12GB VRAM. SIFT keypoints generated from the KITTI dataset are stored using about 400GB of HDD. We only generated SIFT patches of size 15x15 and it took around 40 hours on Intel(R) Core(TM) i7-8700K CPU @ 3.70GHz. For larger sizes, it takes even more time. Hard disk reads are the implementation bottleneck which prevented us from running more than 20 training epochs of batch size 4 in the learning experiments. There are multiple ways to cope with the latter problem. One is to generate compressed SIFT descriptors to reduce the overall dataset size. Another way is to use a GPU implementation of SIFT compiled as PyTorch node and the SIFT keypoints can then be calculated on-demand. However, the compression takes even more time to generate the descriptors. We therefore aim for the second improvement.

\subsection{Other Experiments}

An interesting question is whether the keypoint similarity loss could be used alone, with all the other loss function components' weights set to zero. We ran the experiment with and without the smooth loss component. When run with the smooth loss component, the produced depth maps after an epoch of learning are completely smooth, which means that the model have converged too early. In this case, the smoothness loss was too dominant over the keypoint similarity loss, which might be mitigated by lowering the smooth weight hyperparameter. However, when the smoothness loss was completely turned off, the model was unable to significantly move from the starting point. All this suggests that the keypoint similarity loss in this form can not be solely used as a loss function. Surprisingly, multi-scale depth estimation did not outperform its single-scale counterpart so we do not mention it in the analysis.

\section{\uppercase{Future work}}
    \label{section:future-work}
    
The mandatory step towards reaching state-of-the-art results is further model fine-tuning. It is also worth examining SIFT's parameters further. The most important parameter for the experiments is the size of the keypoint. By using larger patches, low or repeating texture image regions should be learned better. In the experiments, it is shown that the keypoint guided self-supervision replaces the explainability mask. Larger keypoint patches might also serve as an elegant replacement for multi-scale depth map estimations. Ideally, we will wrap SIFT as a computational node, which would enable SIFT to choose the size an orientation of keypoints itself.

However, all these approaches largely depend on, and are limited by, the SIFT performance. Taking a step further, it seems reasonable to try a joint learning of image keypoints and depth estimation in a similar, self-supervised learning fashion.

\subsection{Learning the Keypoints}

Encouraged by LF-Net presented in \cite{lf-net}, we plan to learn the keypoints instead of directly using the ones selected and described by SIFT. The keypoints learning model should implicitly capture the depth properties of a scene. We propose the following high-level pipeline:

\begin{itemize}
    \item Select and describe N SIFT keypoints for every \textit{source} image.
    \item Use the \textit{keypoints network} to find $N$ keypoints on the target image.
    \item Reproject the estimated target image keypoints to the source images based on depth (and pose) estimation.
    \item \textit{Assign} the correspondent source image keypoint to every reprojected target keypoint reprojected and compare them.
\end{itemize}

Note that the correspondences now need to be determined, because they not known upfront. This makes the task much harder, so we decide to simplify it by using the pose ground truths. The dataset in which the camera parameters are known is, for example, an InteriorNet by \cite{interior-net}. The subset of structure-from-motion problems where the camera parameters are known is called a multi-view-stereo or MVS \cite{mvs-tutorial} and it is a possible future direction for joint learning of geometry and keypoints.

\section{\uppercase{Conclusion}}
    \label{section:conclusion}

The aim of our work is to propose a step towards keypoint guided learning of depth estimation. The model performance is still not on the state-of-the-art level on KITTI dataset, but the experimental results suggest that the presented approach improves the learning. The future goals are to wrap SIFT as a PyTorch node and try learning joint keypoint and depth estimation, completely removing the SIFT dependency. To make the joint learning of depth and keypoints easier, we propose to learn under camera parameters semi-supervision, i.e., in a multi-view-stereo configuration.

\section*{\uppercase{Acknowledgement}}

This work has been supported by Croatian Science Foundation under the grant number HRZZ-IP-2018-01-8118 (STEAM) and by European Regional Development Fund under the grant number KK.01.1.1.01.0009 (DATACROSS).

\bibliographystyle{apalike}
{\small
\bibliography{main}
}

\end{document}